\documentclass[twocolumn,letterpaper]{IEEEAerospaceCLS}  % only supports two-column, letterpaper format

\usepackage[]{graphicx}    % We use this package in this document
\newcommand{\ignore}[1]{}  % {} empty inside = %% comment

\usepackage{amsmath,amssymb}
\usepackage{subcaption}
\usepackage{xcolor}
\usepackage{stfloats}

\usepackage{algorithm}
\usepackage{algpseudocode}
\algnewcommand\algorithmicgiven{\textbf{Given:}}
\algnewcommand\Given{\item[\algorithmicgiven]}
\algnewcommand\algorithmiconline{\textbf{Online}}
\algnewcommand\Online{\item[\algorithmiconline]}
\algnewcommand\algorithmicoffline{\textbf{Offline}}
\algnewcommand\Offline{\item[\algorithmicoffline]}
\algnewcommand\algorithmichlmdp{\textbf{High-Level MDP (Offline)}}
\algnewcommand\HLMDP{\item[\algorithmichlmdp]}
\algnewcommand\algorithmicllmdp{\textbf{Low-Level MDPs (Offline)}}
\algnewcommand\LLMDP{\item[\algorithmicllmdp]}
\algnewcommand\algorithmicplan{\textbf{Traverse Planning (Online)}}
\algnewcommand\PLAN{\item[\algorithmicplan]}

\usepackage{bm}
\usepackage{hyperref}
\usepackage{enumitem}
\newlist{inlinelist}{enumerate*}{1}
\setlist*[inlinelist,1]{%
      label=(\arabic*),
  }

\DeclareMathOperator*{\argmax}{arg\,max}

\newcommand{\MDP}{{\mathcal{M}}}
\newcommand{\States}{{\mathcal{S}}}
\newcommand{\state}{s}
\newcommand{\statep}{s'}
\newcommand{\Actions}{{\mathcal{A}}}
\newcommand{\action}{a}
\newcommand{\Transitions}{{\mathcal{T}}}
\newcommand{\Rewards}{{\mathcal{R}}}
\newcommand{\reward}{r}
\newcommand{\policy}{\pi}
\newcommand{\optpolicy}{\pi^*}
\newcommand{\E}{{\mathbb{E}}}

\newcommand{\MDPHL}{{\mathcal{M}^\text{HL}}}
\newcommand{\StatesHL}{{\mathcal{S}^\text{HL}}}
\newcommand{\stateHL}{s^\text{HL}}
\newcommand{\ActionsHL}{{\mathcal{A}^\text{HL}}}
\newcommand{\actionHL}{a^\text{HL}}
\newcommand{\TransitionsHL}{{\mathcal{T}^\text{HL}}}
\newcommand{\RewardsHL}{{\mathcal{R}^\text{HL}}}

\newcommand{\optpolicyHL}{{\pi^*_\text{HL}}}

\newcommand{\MDPLL}{{\mathcal{M}^\text{LL}_\text{i}}}
\newcommand{\StatesLL}{{\mathcal{S}^\text{LL}}}
\newcommand{\stateLL}{s^\text{LL}}
\newcommand{\ActionsLL}{{\mathcal{A}^\text{LL}}}
\newcommand{\actionLL}{a^\text{LL}}
\newcommand{\TransitionsLL}{{\mathcal{T}^\text{LL}}}
\newcommand{\RewardsLL}{{\mathcal{R}^\text{LL}}}
\newcommand{\rewardLL}{{r^\text{LL}}}
\newcommand{\optpolicyLL}{{\pi^*_\text{LL,i}}}
\newcommand{\tgtLL}{{\text{tgt}^\text{LL}}}

\newlength{\picwidth}
\begin{document}
\title{Contingency Planning Using Bi-level Markov Decision Processes for Space Missions}

\author{%
Somrita Banerjee\\ 
Department of Aeronautics and Astronautics\\
Stanford University\\
Stanford, CA 94305\\
somrita@stanford.edu
\and 
Edward Balaban\\
NASA Ames Research Center\\
269 Parsons Ave\\
Moffett Field, CA 94035\\
edward.balaban@nasa.gov
\and 
Mark Shirley\\
NASA Ames Research Center\\
269 Parsons Ave\\
Moffett Field, CA 94035\\
mark.h.shirley@nasa.gov
\and 
Kevin Bradner\\
NASA Ames Research Center\\
269 Parsons Ave\\
Moffett Field, CA 94035\\
kevin.bradner@nasa.gov
\and
Marco Pavone\\ 
Department of Aeronautics and Astronautics\\
Stanford University\\
Stanford, CA 94305\\
pavone@stanford.edu
%%%% IMPORTANT: Use the correct copyright information--IEEE, Crown, or U.S. government. %%%%%
\thanks{\footnotesize 979-8-3503-0462-6/24/$\$31.00$ \copyright2024 IEEE}              % This creates the copyright info that is the correct 2024 data.
%\thanks{{U.S. Government work not protected by U.S. copyright}}         % Use this copyright notice only if you are employed by the U.S. Government.
%\thanks{{979-8-3503-0462-6/24/$\$31.00$ \copyright2024 Crown}}          % Use this copyright notice only if you are employed by a crown government (e.g., Canada, UK, Australia).
%\thanks{{979-8-3503-0462-6/24/$\$31.00$ \copyright2024 European Union}}    % Use this copyright notice is you are employed by the European Union.
}

\maketitle

\thispagestyle{plain}
\pagestyle{plain}

\begin{abstract}
This work focuses on autonomous contingency planning for scientific missions by enabling rapid policy computation from any off-nominal point in the state space in the event of a delay or deviation from the nominal mission plan. Successful contingency planning involves managing risks and rewards, often probabilistically associated with actions, in stochastic scenarios. Markov Decision Processes (MDPs) are used to mathematically model decision-making in such scenarios. However, in the specific case of planetary rover traverse planning, the vast action space and long planning time horizon pose computational challenges. A bi-level MDP framework is proposed to improve computational tractability, while also aligning with existing mission planning practices and enhancing explainability and trustworthiness of AI-driven solutions. We discuss the conversion of a mission planning MDP into a bi-level MDP, and test the framework on RoverGridWorld, a modified GridWorld environment for rover mission planning. We demonstrate the computational tractability and near-optimal policies achievable with the bi-level MDP approach, highlighting the trade-offs between compute time and policy optimality as the problem's complexity grows. This work facilitates more efficient and flexible contingency planning in the context of scientific missions.
\end{abstract}

\tableofcontents

%%%%%%%%%%%%%%%%%%%%%%%%%%%%%%%%%%%%%%
\section{Introduction}
%%%%%%%%%%%%%%%%%%%%%%%%%%%%%%%%%%%%%%

\begin{figure*}[htb]
    \centering
    \includegraphics[width=0.9\textwidth]{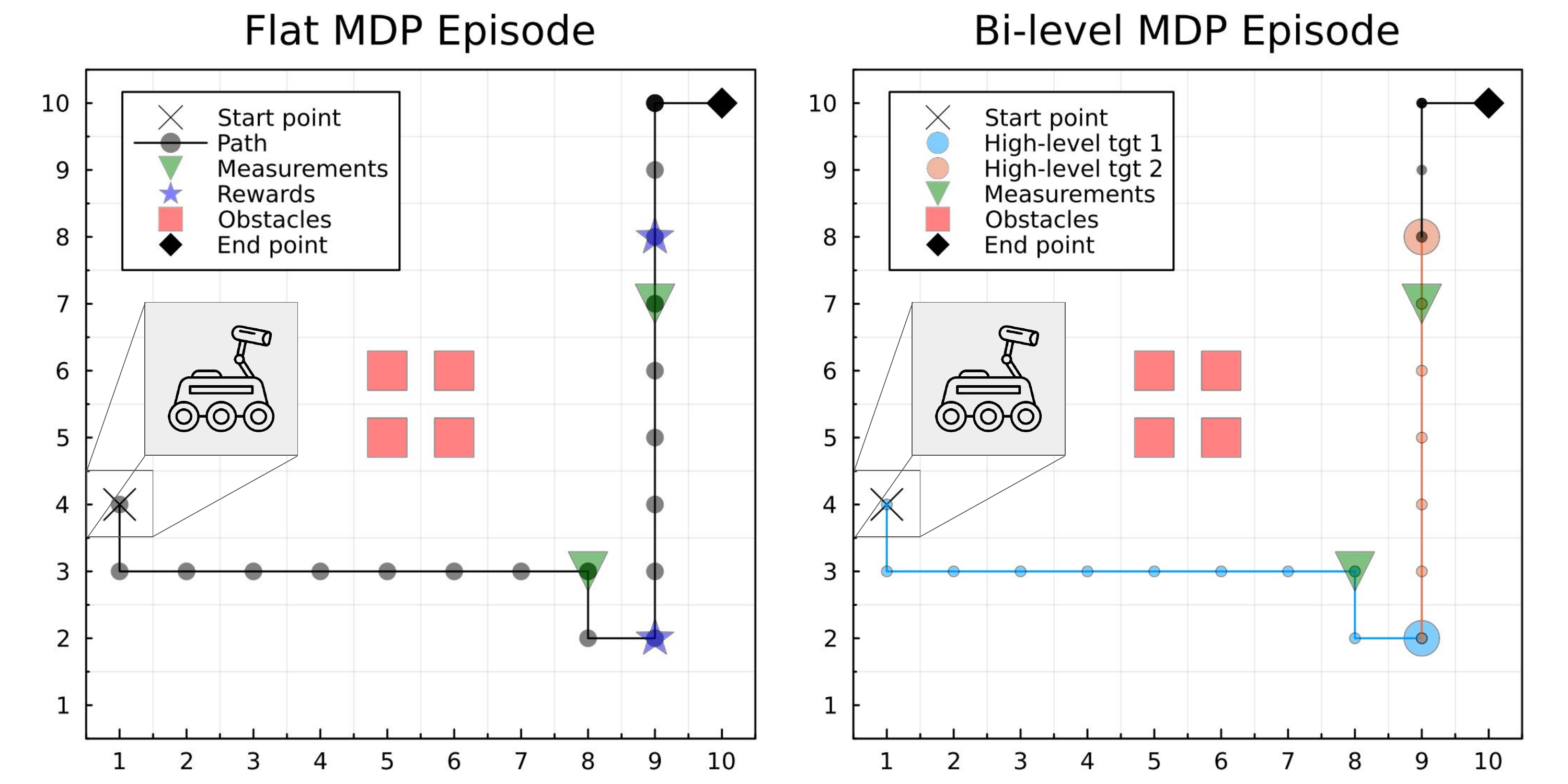}
    \caption{\bf{A rover path planning problem to collect measurements and drill at targets, while avoiding obstacles and moving sun shadows (not pictured), is solved using a flat MDP and a bi-level MDP. In the bi-level MDP formulation, the high-level MDP decides which scientific target to drill next, while the low-level MDP plans the path and decides placement of measurements. Each target can only be drilled once, and must be preceded by a measurement at a neighboring cell (including diagonal cells). In both formulations, the optimal path is the same. However, solving the bi-level MDP is faster. }}
    \label{fig:exp2-episodes}
\end{figure*}

%% Motivation
There is a growing need for greater autonomy in scientific mission planning, precipitated by the rise in the number, duration, and complexity of missions. These scientific missions are inherently stochastic, and risk/reward management is a fundamental aspect of successful mission planning. In this work, we are motivated by NASA’s Volatiles Investigating Polar Exploration Rover (VIPER) and its planned mission to the lunar South Pole \cite{ColapreteAndrewsEtAl2019,HeldmannColapreteEtAl2016}. Like any scientific mission, this mission involves balancing multiple considerations such as time or cost constraints, as well as risks and scientific rewards associated, often probabilistically, with possible activities. Such decision-making in stochastic scenarios can be mathematically modeled through a framework known as Markov Decision Processes (MDPs) \cite{Bellman1957}. An MDP is characterized by a set of states, a set of actions, a probabilistic transition function from the current state to the next state, and a reward function mapping each potential action at a system state to a reward value. 

However, in the case of scientific mission planning, especially for traverse planning for planetary rovers, the action space can be very large and the planning time horizon can be many orders of magnitude greater than the granular time steps at which actions must be taken. For example, at each step of a traverse, a lunar rover must decide which action to take -- it may move in multiple directions, with varying speeds depending on terrain traversability, azimuth to the sun, and current science objectives, while making progress towards the goal target for the next area of scientific interest or location favorable for waiting out the lunar night. If these decisions are made, for instance, every hour, while accumulating rewards over the total mission lifetime (e.g., over four Earth months in the case of VIPER), that would yield a horizon of over 100,000 timesteps. It is therefore computationally expensive to generate the optimal policy, i.e., the mapping from each state to the optimal action to take to maximize expected future rewards. Furthermore, while a \textit{policy} is the mathematically sound way of acting in the presence of uncertainty, in practice we do need to find a comfortable middle ground with existing operational paradigms. Specifically, this means generating a nominal \textit{plan} (i.e., a series of sequential states the rover is likely to follow), and contingency branches for off-nominal scenarios. This is often done by computing an optimal or near-optimal policy, determinizing it using the maximum likelihood state transitions and rewards to produce a robust plan, and then selecting critical points on that plan to generate contingency branches.

We are inspired by a view of contingency planning in which an optimal or near-optimal policy is calculated from any off-nominal point in the state space (e.g., any geographical position, at any time, any battery state-of-charge), such that each such policy forms a contingency branch. In this view of contingency planning, we consider off-nominal states that are caused by delays relative to or deviations from the nominal plan, or fault events (e.g., to the measurement instruments), but without any change in driving characteristics of the rover. These policies can be generated offline prior to the mission and re-generated with updated state information during the mission. The computation of these policies can quickly become intractable, if solved using a fine-grained MDP. Motivated by this view of contingency planning, we focus on finding a tractable approach to quickly compute near-optimal policies from any off-nominal state, a problem that also generalizes to nominal traverse planning.

To combat the intractability of a flat fine-grained MDP, we introduce a bi-level MDP formulation for traverse planning. A bi-level framework is naturally suited, and already widely-used, in mission planning. Using VIPER as an example, at a high-level, the strategic plan decides which scientific areas of interest (referred to as science stations) to target and where the rover should hibernate through the lunar night (to minimize communication loss and sun shadows). At a low level, the tactical planner computes the fine-grained actions to drive to the specified target and measurements to take along the way. A bi-level MDP formalizes this structure to enable fast and autonomous computation of optimal policies. The bi-level structure also allows for introduction of human preferences and makes the decisions explainable and investigable, increasing acceptance and trustworthiness of AI solutions. 

%% Related work
The use of MDPs to formulate path planning problems has been widely explored. MDPs have been used for path planning for unmanned aerial vehicles \cite{Al-SabbanGonzalezEtAl2013,AllamarajuKingraviEtAl2014}, autonomous marine vehicles in ocean flows \cite{ChowdhurySubramani2022,KularatneHajiegharyEtAl2018}, urban navigation robots \cite{NardiStachniss2019}, and even flexible needles for minimally invasive surgery \cite{TanYuEtAl2018}. MDPs have been used for rover traverse planning \cite{OnoRothrockEtAl2020,YuWangEtAl2021} but have necessarily used a pared-down state and action space or a short planning horizon, with a common concern mentioned in the literature being the loss of tractability as the state and action spaces increase in size \cite{FolsomOnoEtAl2021}.  

% Quote from another Ono paper: (https://journals.sagepub.com/doi/full/10.1177/1729881421999587) A similar Mars rover navigation problem was solved using an MDP formulation; however, this research only considered slope when evaluating traversability, assumed a coarsse 20 × 20 m2 cell size, and did not consider a cooperating helicopter.10 The addition of terrain type and rock abundance to their model could significantly increase the state and observation spaces.

To increase tractability as state spaces and time horizons get larger, various frameworks have been developed for planning using \textit{hierarchical} MDPs. Frameworks within the field of hierarchical reinforcement learning (such as MAXQ \cite{Dietterich2000}) increase tractability but are model-free and require millions of interactions with the environment to learn a task. Model-based frameworks, such as the hierarchical MDP framework for robot planning by Bakker et al. \cite{BakkerZivkovicEtAl2005}, both increase tractability and leverage existing robot models, thus inspiring the framework presented in this paper. Our approach has some key differences, namely that (1) our framework is targeted towards mission planning, i.e., it can generate a bi-level MDP from a mission planning MDP along with a few pieces of mission-specific information, without the use of state clustering functions or averaging over state transition probabilities and rewards (2) our framework does not use nested value iteration but instead supplants the high-level transition function with a heuristic whose fidelity can be tuned to make a trade-off between optimality and computation time and (3) our framework ties the idea of MDP target states to the mission planning goal states, and shares the same target across all levels of the MDP. Nevertheless, this work shares with the framework presented in \cite{BakkerZivkovicEtAl2005}, the ability to reduce the state and action space size to increase computational tractability. Such hierarchical MDPs, specifically bi-level MDPs, have been used successfully in other domains, e.g., to balance electric vehicle charging load with the availability of wind power \cite{LongMaEtAl2019}. The domain of space mission planning is already well-suited to a bi-level planning structure, where the high-level planner decides the placement of activities and the low-level planner executes those activities based on an activity dictionary \cite{Ai-ChangBresinaEtAl2004,EstlinFisherEtAl2002}.

This paper aims to introduce bi-level MDPs specifically for scientific mission planning, by describing how key decision points can be chosen to create a bi-level framework, and demonstrating the resulting improvement in tractability, which unlocks the ability to generate traverse plans from any arbitrary state and can be used to form contingency branches. An example plan from an arbitrary state is shown in Figure \ref{fig:exp2-episodes} and discussed further in Section \ref{sec:results}.

The contributions of this paper are as follows: We introduce a modified GridWorld environment that captures the time-sensitive constraints in rover mission planning, which we call RoverGridWorld. We present a framework for handling mission planning with large action spaces and planning horizons. We show how a mission planning MDP can be converted into a bi-level MDP by identifying mission-relevant decision points (or branching-off points). We compare our bi-level MDP approach to the original flat MDP, and show that our approach is more computationally tractable while achieving near-optimal policies. As the complexity of the problem increases, i.e., action spaces and state spaces get larger and time horizons get longer, we examine the trade-offs between computational complexity and optimality of the policy.

The paper is organized as follows. Section \ref{sec:background} briefly describes the MDP framework and methods for solving an MDP. Section \ref{sec:problemformulation} describes the RoverGridWorld problem and its representation of the mission planning problem. Section \ref{sec:approach} describes how a flat MDP is converted to a bi-level MDP, and the additional inputs required by the mission planner. Section \ref{sec:results} presents results of comparing the bi-level MDP approach to other solvers, and demonstrates the computational advantages offered by the bi-level framework, i.e., achieving 78\% of the optimal reward using 13\% of the computation time. Finally, Section \ref{sec:conclusion} provides conclusions and future research directions.

%%%%%%%%%%%%%%%%%%%%%%%%%%%%%%%%%%%%%%
\section{Background}
\label{sec:background}
%%%%%%%%%%%%%%%%%%%%%%%%%%%%%%%%%%%%%%
The task of path planning in potentially stochastic scenarios is often formalized as a Markov Decision Process (MDP). This framework has several advantages: (1) it can account for uncertainty in state transitions, e.g., when an action has stochastic effects, (2) policies can be computed for the entire state space, if offline solvers are used, allowing for fast online lookup of the optimal actions from any given state, (3) the policies maximize expected cumulative reward, over the probability space of transitions, and (4) the framework allows for easy inclusion of a custom rewards function, which, for path planning, means taking into account targets, obstacles, and constraints. Simple MDPs are often solved exactly using dynamic programming methods, such as value iteration. We introduce these concepts briefly.
\subsection{Markov Decision Processes (MDPs)}
An MDP is a problem formulation for sequential decision making under uncertainty and is represented as $\MDP : \langle \States, \Actions, \Transitions, \Rewards \rangle$. The state space $\States$ is the set of states $\state$ that the agent can be in, including terminal states. The action space $\Actions$ is the set of all possible actions $\action$ that the agent may take. The transition function $\Transitions : \States \times \Actions \times \States \rightarrow [0,1]$ describes the probability that taking action $\action$ from state $\state$ will lead to state $\statep$, i.e., $\Transitions\left(\state, \action, \statep \right) = p\left( \state_{t+1} = \statep \mid \state_t = \state, \action_t = \action \right)$. To satisfy the Markov property, the probability that the agent moves to the next state $\statep$ at time $t+1$ only depends on the current state $\state_t$ and current action $\action_t$, and is conditionally independent of all previous states and actions \cite{Bellman1957}. The reward function $\Rewards : \States \times \Actions \times \States \rightarrow \mathbb{R}$ defines the immediate reward $r\left(\state, \action, \statep \right)$ received by the agent when action $\action$ is taken from state $\state$ leading to state $\statep$.

A policy is defined as a mapping $\policy : \States \rightarrow \Actions$, and a policy execution requires that when the agent is in state $\state$, the agent takes the action $\policy(\state)$. Solving an MDP is synonymous with finding an optimal policy $\optpolicy$ that maximizes the expected, possibly discounted, future cumulative reward. The episode ends when the agent reaches a terminal state (e.g., in the rover planning case, because a pre-specified terminal goal is reached or the mission viability has ended). This corresponds to maximizing the estimated return from each state, which is captured in a value function $V$. The optimal policy therefore yields the optimal value function, i.e., 
\begin{equation}
\label{eqn:value-function}
    V^*(\state) = \max_\policy \E \left(\sum_{t=0}^T \gamma^t \reward(\state_t, \policy(\state_t), \state_{t+1}) \mid \policy, \state_0 = s \right)
\end{equation}
for each state $\state$, maximum time horizon $T$, and where the discount factor $\gamma$ is a parameter used to balance the relative importance of earlier and later rewards. The expectation operator $\E$ averages over rewards and stochastic state transitions.

\subsection{Solving using Value Iteration}
MDPs where the model is known, i.e., with known transition and reward functions, can be solved optimally through dynamic programming \cite{Bertsekas1995}. Dynamic programming iteratively calculates the value function $V(\state)$, and is guaranteed to converge to the optimal value function $V^*(\state)$ as described in Equation \ref{eqn:value-function}. In this work, we focus on one dynamic programming approach, value iteration, which updates the value function for every state in the state space, i.e., 
\begin{equation}
    V(\state) \leftarrow \max_{\action} \sum_{\statep} p(\statep \mid \state, \action) \left[\reward(\state, \action, \statep) + \gamma V(\statep) \right].
\end{equation}
The value functions are updated iteratively until the change in value function between iterations becomes small \cite{Bertsekas1995}. Once the value function for each state $V(\state)$ has converged to $V^*(\state)$, the optimal policy $\optpolicy$ is derived by taking the action that maximizes the value function at each state:
\begin{equation}
    \optpolicy(\state) = \argmax_{\action} \sum_{\statep} p(\statep \mid \state, \action) \left[\reward(\state, \action, \statep) + \gamma V^*(\statep) \right]
\end{equation}

\subsection{Solving using Reinforcement Learning (RL)}
For most of this paper, we will be using value iteration to solve the MDP, since the transition and reward functions are known. However, we will also evaluate results using two solvers from the reinforcement learning family, which solve MDPs without knowledge of the transition probabilities. These algorithms, such as Q-learning and SARSA, gather experience of the environment through simulations and use that to update an internal representation of experience, known as the Q value:
\begin{equation}
    Q(\state, \action) \leftarrow \sum_{\statep} p(\statep \mid \state, \action) \left[\reward(\state, \action, \statep) + \gamma V(\statep) \right].
\end{equation}
This representation, refined through simulations, is used to generate a policy. While these RL solvers are useful for real problems with unknown transitions and large state spaces, in this work we focus on demonstrating the value of our bi-level MDP on problems with known transitions and rewards. In the next section, we describe our problem formulation.

%%%%%%%%%%%%%%%%%%%%%%%%%%%%%%%%%%%%%%
\section{Problem Formulation}
\label{sec:problemformulation}
%%%%%%%%%%%%%%%%%%%%%%%%%%%%%%%%%%%%%%
Inspired by mission planning for a rover mission such as VIPER, we formulate our problem as a modified version of GridWorld. The modifications are introduced to better capture the challenges faced during rover mission planning, so we refer to the problem as RoverGridWorld \footnote{The code for RoverGridWorld and our bi-level MDP formulation can be found at  \url{https://github.com/somritabanerjee/BiMDPs.jl}.}. Some key features of the RoverGridWorld are:
\begin{itemize}
    \item Actions are either movements or in-place activities (measurement or drilling). Given a movement action, the transition function always moves the rover to the intended direction with no stochasticity.
    \item Time is an important component of the state. Many parts of the mission rely on being at the right place at the right time, e.g., a target can only be visited when it is not in communication shadow. 
    \item The state also tracks which targets the rover has visited, to prevent repetition. To simulate science constraints, the rover is required to take a measurement near each target before that target can be drilled.
    \item The rewards function is time-dependent, i.e., certain rewards for targets are only available for a given time window. Some obstacles are dynamic as well, e.g., to simulate sun shadows moving across the lunar polar region (the rover is assumed to be solar-powered, therefore driving through sun shadows is undesirable).
\end{itemize}

The RoverGridWorld MDP can be denoted as 
\begin{equation*}
    \MDP : \langle \States, \Actions, \Transitions, \Rewards \rangle
\end{equation*}
where the state tracks current position, time, and target history, i.e.,
\begin{equation*}
\States = \{\state\} = \{\left(x, y, t, \texttt{measured}, \texttt{drilled} \right)\}
\end{equation*}
such that $\texttt{measured}[i] = $\textit{true} if the $i$th target has an associated measurement (conducted in any neighboring cell) and \textit{false} if not, and $\texttt{drilled}[i] = $\textit{true} if the $i$th target has been drilled and \textit{false} if not. The actions are either movements or in-place activities, i.e.,
\begin{equation*}
\Actions = \{\texttt{UP, DOWN, LEFT, RIGHT, MEASURE, DRILL}\}.
\end{equation*}
The transition function $\Transitions(\state, \action)$ moves the rover to the correct grid location if available, and stays in place if the movement is not possible or if a measurement/drilling action is being conducted. The transition function also advances time for the states and tracks which targets have been measured or drilled. The transition function transitions to the terminal state if the rover has reached one of the pre-specified goals. The reward function $\Rewards(\state, \action)$ rewards measuring and drilling various targets and also penalizes hitting a static obstacle or getting caught in time-varying sun shadows, i.e., 
\begin{equation*}
\Rewards(\state, \action) = \Rewards_\text{tgts}(\state, \action) + \Rewards_\text{obst}(\state, \action).
\end{equation*}
A reward for drilling a target is only given if a measurement has been conducted in a cell neighboring the target (including diagonal cells).

\begin{figure*}
     \centering
    \begin{tabular}[c]{cccc}
     \begin{subfigure}[b]{\picwidth}
         \centering
         \includegraphics[width=\textwidth]{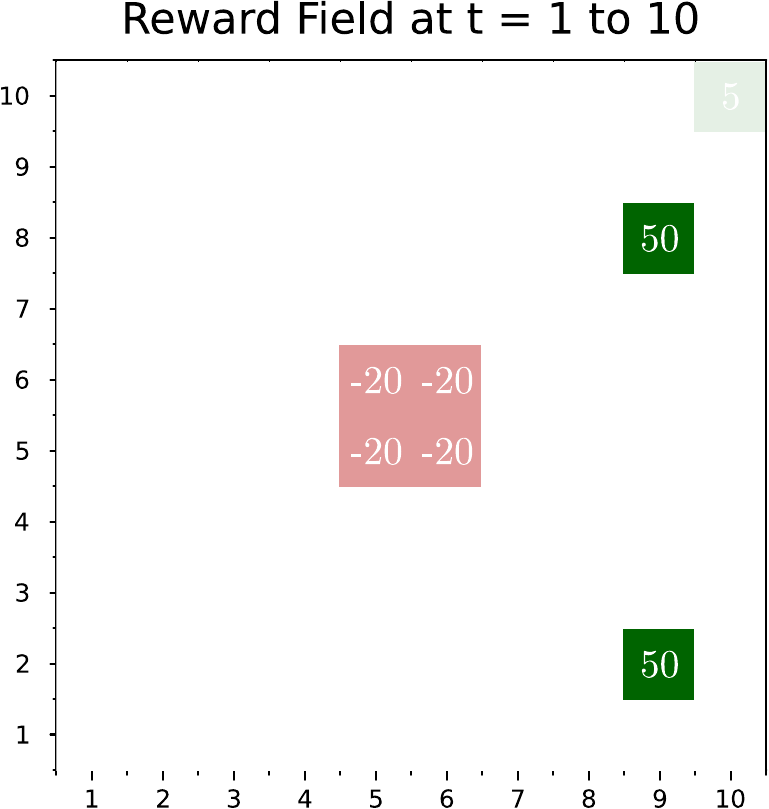}
     \end{subfigure}&
     \hfill
     \begin{subfigure}[b]{\picwidth}
         \centering
         \includegraphics[width=\textwidth]{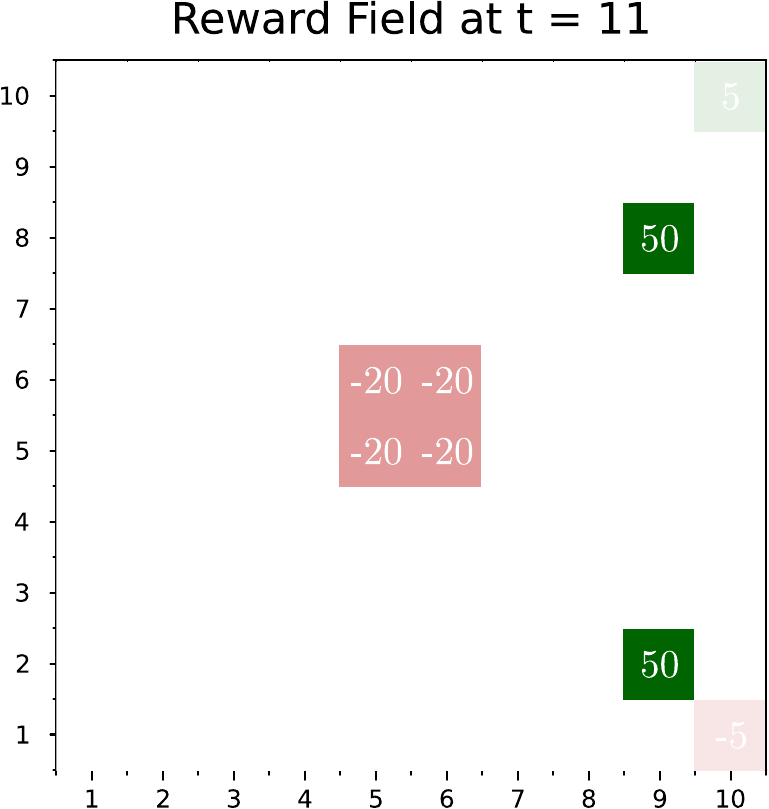}
     \end{subfigure}&
     \hfill
     \begin{subfigure}[b]{\picwidth}
         \centering
         \includegraphics[width=\textwidth]{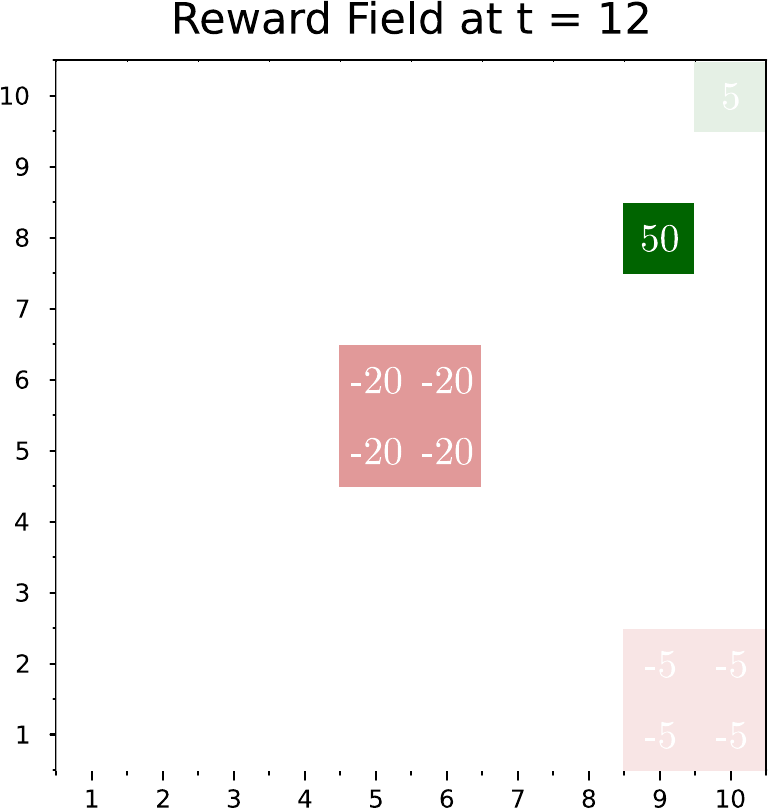}
     \end{subfigure}&
     \hfill
     \begin{subfigure}[b]{\picwidth}
         \centering
         \includegraphics[width=\textwidth]{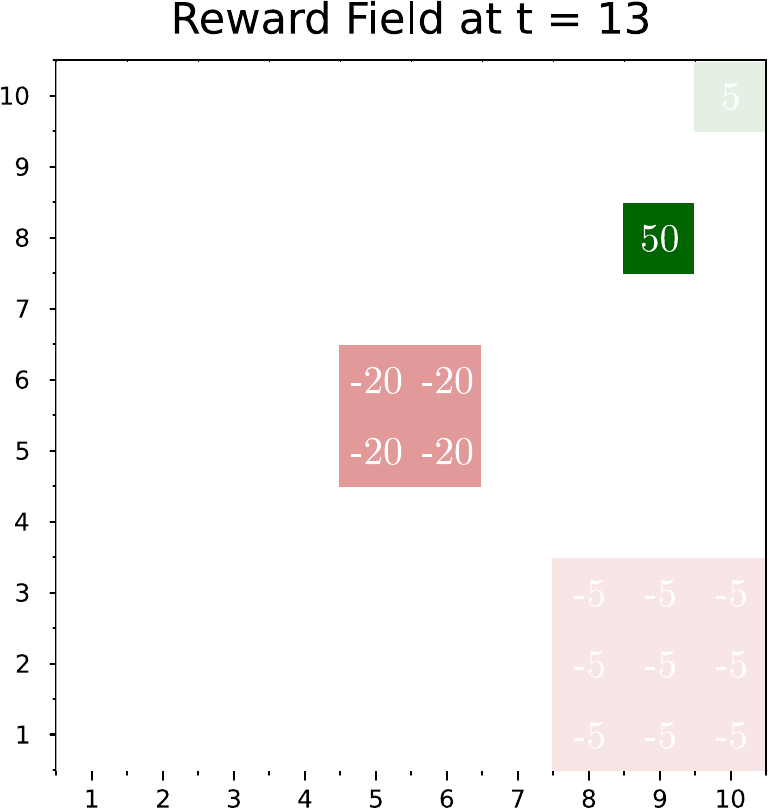}
     \end{subfigure}\\
     \hfill
     \begin{subfigure}[b]{\picwidth}
         \centering
         \includegraphics[width=\textwidth]{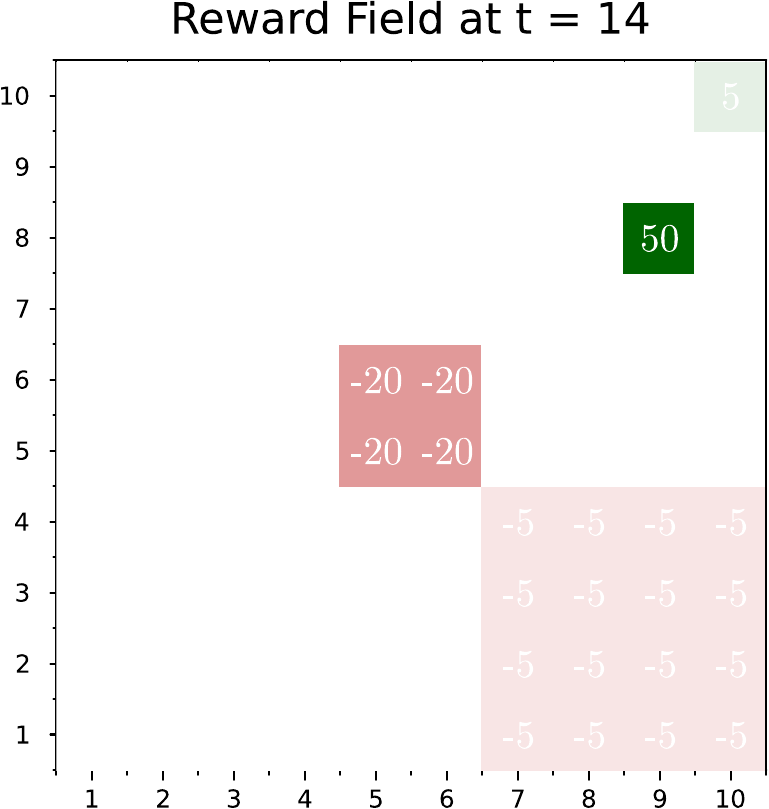}
     \end{subfigure}&
     \hfill
     \begin{subfigure}[b]{\picwidth}
         \centering
         \includegraphics[width=\textwidth]{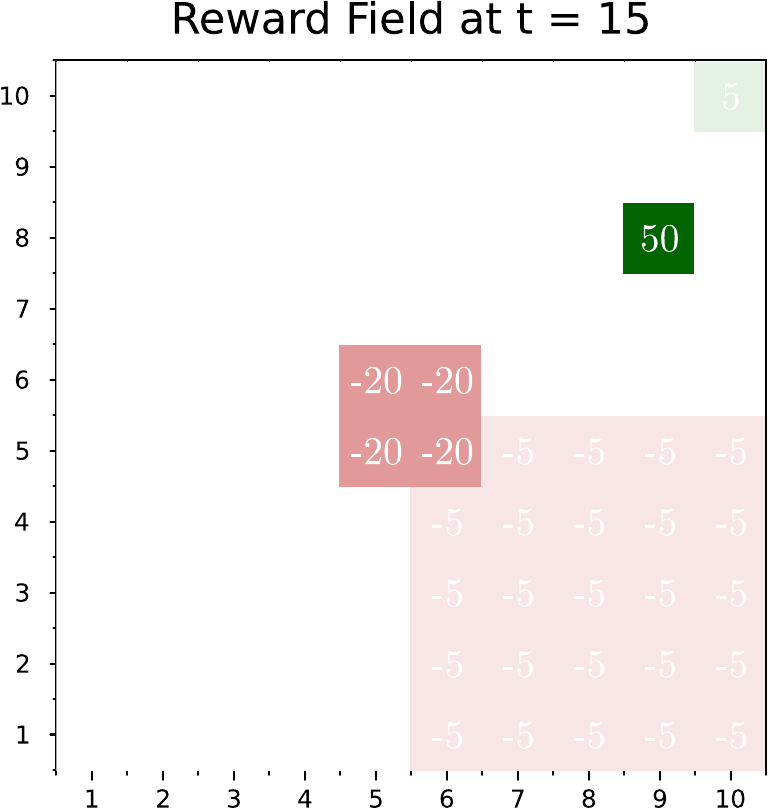}
     \end{subfigure}&
     \hfill
     \begin{subfigure}[b]{\picwidth}
         \centering
         \includegraphics[width=\textwidth]{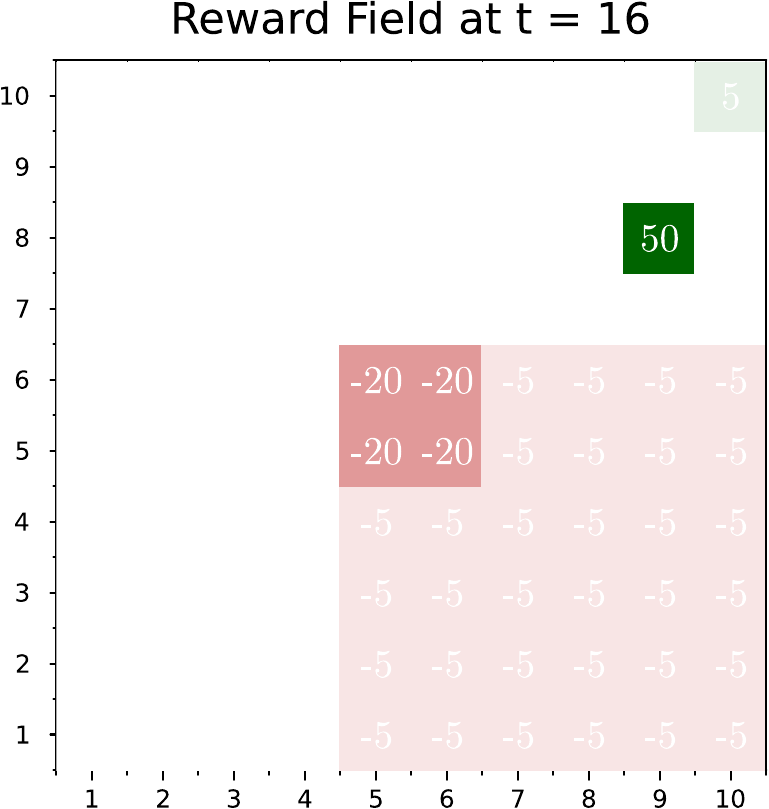}
     \end{subfigure}&
     \hfill
     \begin{subfigure}[b]{\picwidth}
         \centering
         \includegraphics[width=\textwidth]{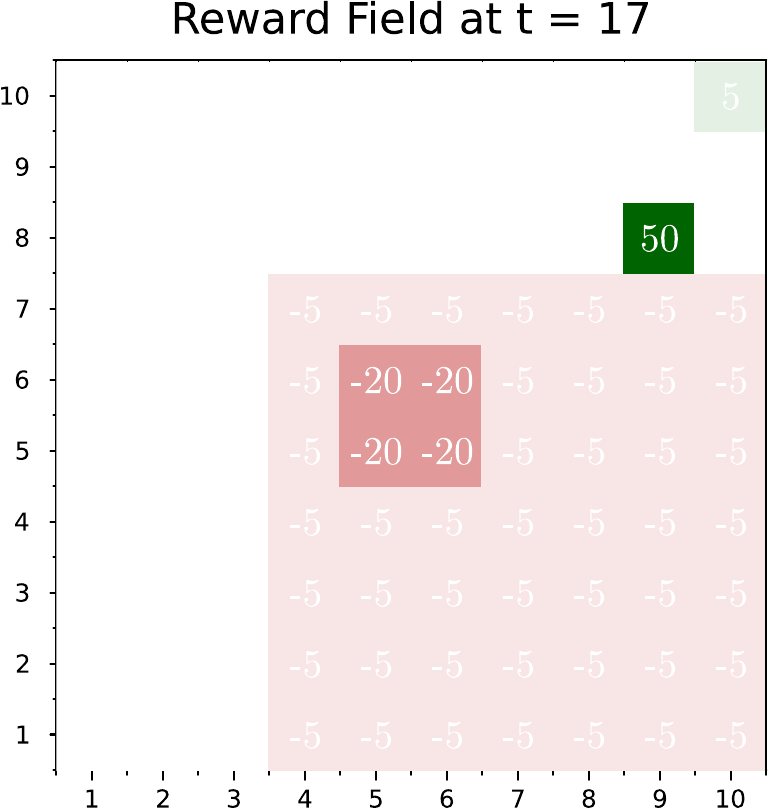}
     \end{subfigure}\\
     \hfill
     \begin{subfigure}[b]{\picwidth}
         \centering
         \includegraphics[width=\textwidth]{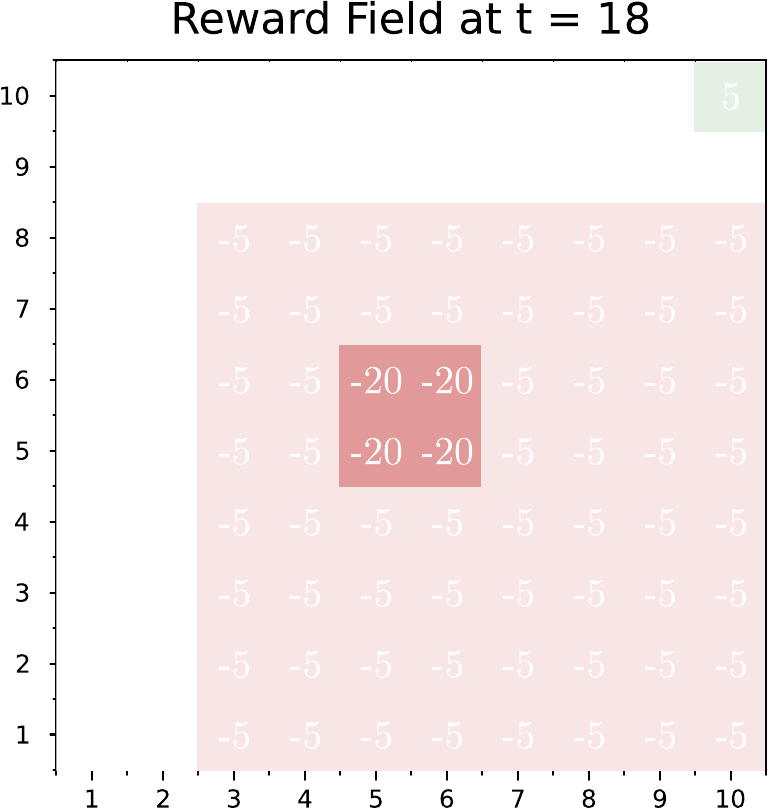}
     \end{subfigure}&
     \hfill
     \begin{subfigure}[b]{\picwidth}
         \centering
         \includegraphics[width=\textwidth]{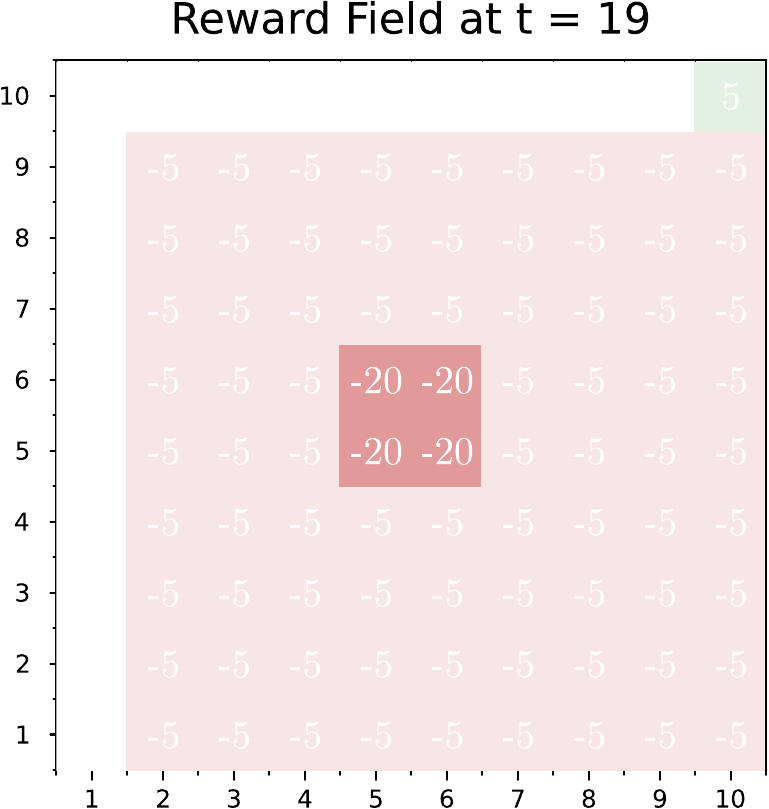}
     \end{subfigure}&
     \hfill
     \begin{subfigure}[b]{\picwidth}
         \centering
         \includegraphics[width=\textwidth]{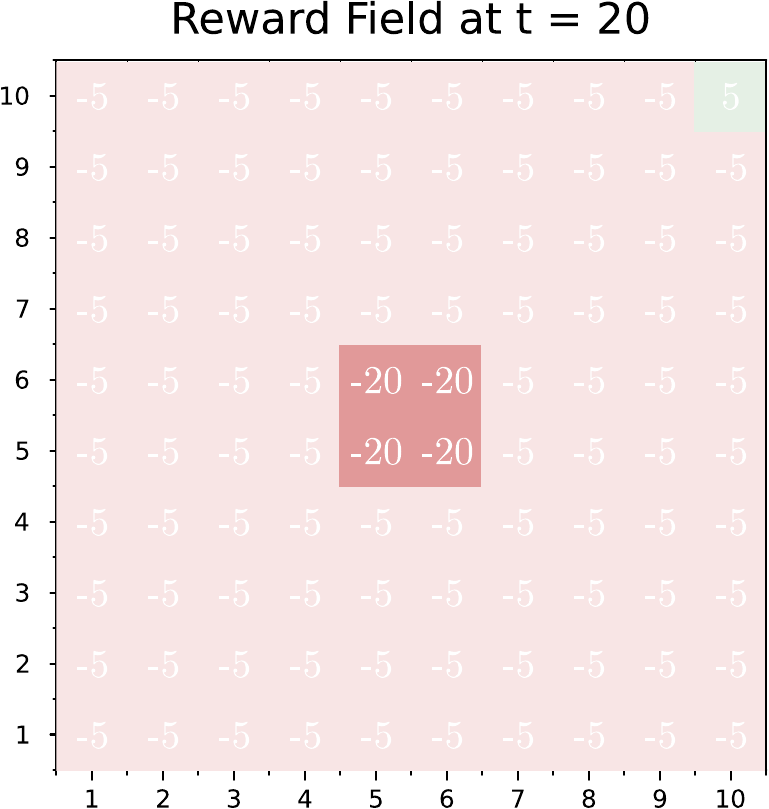}
     \end{subfigure}\\
    \end{tabular} 
    \caption{\bf{A sample reward field for RoverGridWorld that includes targets as well as static and dynamic obstacles. The higher reward targets (dark green) correspond to science stations and the lower reward target (light green) corresponds to a hibernation area. Note that the reward for a hibernation area is only included for ease of visualization, and is not important to the problem. The static obstacles (dark red) are used to model hazardous terrain and the dynamic obstacles (light red) are used to model swiftly moving sun shadows, e.g., those near the lunar polar regions. By the end of the time horizon, if the rover is not at the hibernation area, it begins accruing negative rewards.}}
    \label{fig:rewards-field}
\end{figure*}

The time-varying rewards field for a small RoverGridWorld problem is described in Figure \ref{fig:rewards-field}. This small problem is presented for ease of visualization but the experiments with RoverGridWorld in Section \ref{sec:results} use a large planning horizon compared to the time discretization and use a larger state space, i.e., larger grid and more targets. In the next section, we discuss how problems with large state and action spaces are converted into bi-level MDPs for efficient solving.

%%%%%%%%%%%%%%%%%%%%%%%%%%%%%%%%%%%%%%
\section{Approach}
\label{sec:approach}
%%%%%%%%%%%%%%%%%%%%%%%%%%%%%%%%%%%%%%

\begin{algorithm}[h]
\caption{Bi-level MDP Construction and Solving}\label{alg:cap}
\begin{algorithmic}[1]
\Given{$\MDP^0 : \langle \States, \Actions, \Transitions, \Rewards \rangle$ + tgts of length $\mathcal{N}_\text{tgts}$}
\State \hspace*{2em} $\States = \{\state\} = \{(\state_\text{telemetry}, \state_\text{tracking})\}$ 
\State \hspace*{2em} $\Rewards = \Rewards_\text{obst} + \Rewards_\text{tgts}$ 
\State
\HLMDP
\State Construct $\MDPHL: \langle \StatesHL, \ActionsHL, \TransitionsHL, \RewardsHL \rangle$
\State \hspace*{1em} $\StatesHL = \States$ 
\State \hspace*{1em} $\ActionsHL =$ \{$i$ for $i$ in $1:\mathcal{N}_\text{tgts}$ \} \Comment{\textit{Reduced action space}} \label{line_HL_a}
\State \hspace*{1em} $\TransitionsHL(\state, \actionHL, \statep) =$ heuristic($\state, \actionHL, \statep$) \label{line_HL_transitions}
\State \hspace*{1em} $\RewardsHL = \Rewards_\text{tgts}$ \label{line_HL_rewards}
\State $\optpolicyHL \leftarrow$ Solve($\MDPHL$)
\State
\LLMDP
\For{$\actionHL$ in $\ActionsHL$}
    \State $i \leftarrow \actionHL$
    \State $\tgtLL \leftarrow$ tgts$[i]$
    \State Construct $\MDPLL: \langle \StatesLL, \ActionsLL, \TransitionsLL, \RewardsLL \rangle$
    \State \hspace*{1em} $\StatesLL = \{\stateLL\} = \{\state_\text{telemetry}\}$ \Comment{\textit{Smaller states}} \label{line_LL_s}
    \State \hspace*{1em} $\ActionsLL = \Actions$
    \State \hspace*{1em} $\TransitionsLL= \Transitions$ + terminate if $\state$ == $\tgtLL$ 
    \State \hspace*{1em} $\RewardsLL = \Rewards_\text{obst}$ + reward for $\tgtLL$ \label{line_LL_rewards}
    \State $\optpolicyLL \leftarrow$ Solve($\MDPLL$)
\EndFor
\State
\PLAN
\State $\state \leftarrow \state_0$ or rand($\States$)
\State $\stateHL \leftarrow \state$

\While{not terminal($\stateHL$)}
    \State $\actionHL \leftarrow \optpolicyHL(\stateHL)$
    \State $i \leftarrow \actionHL$
    \State $\tgtLL \leftarrow \text{tgts}[i]$
    \State $\stateLL \leftarrow$ SubsetOf$(\stateHL)$
    \While{not terminal($\stateLL$)}
        \State $\actionLL \leftarrow \optpolicyLL(\stateLL)$
        \State $\reward \leftarrow \rewardLL(\stateLL,\actionLL) + \gamma \cdot \reward$
        \State Add to history $(\stateLL, \actionLL, \rewardLL)$
    \EndWhile
    \State $\stateHL \leftarrow$ UpdateHLState$(\stateLL, \text{history})$
\EndWhile
\State return $\reward,$ history

\end{algorithmic}
\label{blmdp}
\end{algorithm}

Given an MDP for mission planning, we formulate a bi-level MDP. The high-level MDP takes the role of the strategic planner and determines which target to go to next. For each selected target, a low-level MDP is constructed that does the tactical planning of navigating to the selected target. 

In order to automatically split the MDP into two levels, our approach requires a few pieces of user input, specific to mission planning. First, we require that the user specify a list of targets $\texttt{tgts}$, which correspond to the high reward states. For a rover, these could be areas of high scientific interest. Second, we require that the reward function $\Rewards$ can be split up into two parts: (1) $\Rewards_\text{tgts}$ that rewards activities associated with targets, this will form the rewards function for the high-level MDP, and (2) $\Rewards_\text{obst}$ that captures all other rewards or penalties, including those due to static or dynamic obstacles, this will form the majority of the rewards function for the low-level MDP. For a rover, $\Rewards_\text{tgts}$ captures rewards due to taking measurements or drilling at a science station and $\Rewards_\text{obst}$ captures penalties due to traversing difficult terrain or getting caught in a sun shadow. Third, we require that the state $\state$ can be broken down into two components: (1) $\state_\text{telemetry}$ which captures instantaneous information, this will form the state space for the low-level MDP and (2) $\state_\text{tracking}$ which stores a snapshot of previous exploration. For a rover, $\state_\text{telemetry}$ includes the rover's XY position, time, battery state-of-charge, etc. and $\state_\text{tracking}$ stores information about the rover's exploration such as which targets have been measured or drilled at, to avoid repetition. These three pieces of information are likely to already be a part of any mission planning architecture. Given these inputs, the MDP can be re-formulated as a bi-level problem as shown in Algorithm \ref{blmdp}.

The key insights that lead to greater computational efficiency are as follows. The high-level MDP's action space (line \ref{line_HL_a}) is limited to only the selection of one of the targets. The high-level MDP's transition function (line \ref{line_HL_transitions}) is a heuristic for the transition from the current state to the target, and its fidelity can be adjusted to trade-off between optimality and computational speed. For the fastest heuristic, paths can be planned coarsely by ignoring low-penalty obstacles and using a conservative model of rover speed. The heuristic could be made optimal, while sacrificing some computation time, by solving the corresponding low-level MDP exactly for each transition. In practice, we do not pre-compute all the low-level MDPs $\MDPLL \text{for } i \text{ in }1:\mathcal{N}_\text{tgts}$, but instead construct them on-the-fly as needed and store for future use. Further, the low-level MDP's state space (line \ref{line_LL_s}) is only the telemetry part of the full state space. The functions \texttt{SubsetOf} and \texttt{UpdateHLState} convert between low-level and high-level states by paring down the state or adding on tracking information to the state. This allows us to keep the state space small for the low-level MDP. 

Since MDP solver convergence time often depends on the size of the state and action space (e.g., value iteration for an MDP with $q$ states and $m$ possible actions requires at most $\mathcal{O}$($mq$) operations in the deterministic case and $\mathcal{O}$($mq^2$) operations in the stochastic case \cite{Bertsekas1995}), the reduction in state and action space sizes lead to performance improvements of the bi-level MDP formulation over the original flat fine-grained MDP. This speedup comes with a trade-off in optimality, but, as we show in the next section, it is often a small trade-off.

%%%%%%%%%%%%%%%%%%%%%%%%%%%%%%%%%%%%%%
\section{Results}
\label{sec:results}
%%%%%%%%%%%%%%%%%%%%%%%%%%%%%%%%%%%%%%
\subsection{Experiment 1}
\begin{figure*}[ht]
    \centering
    \includegraphics[width=0.9\textwidth]{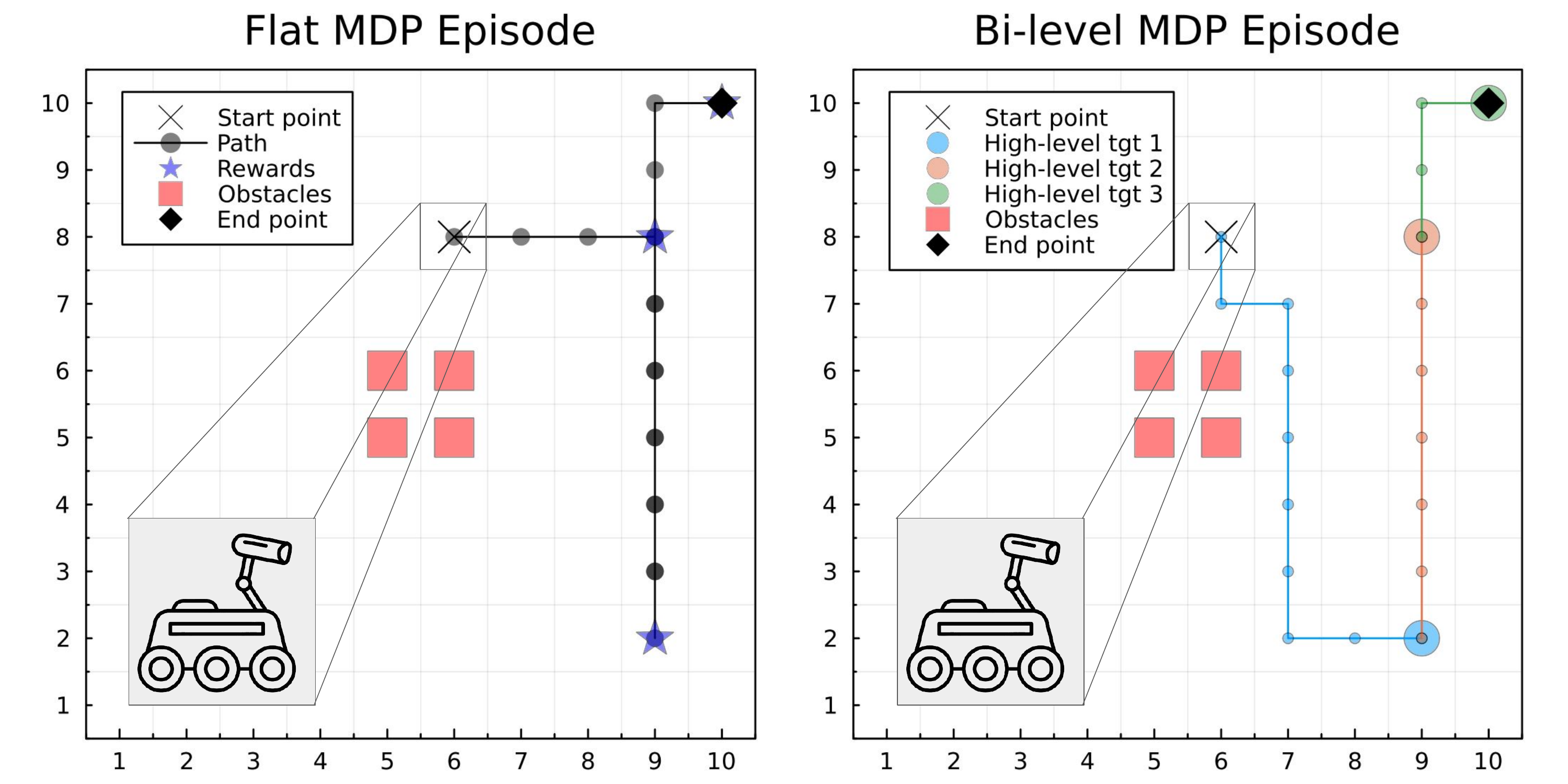}
    \caption{\bf{A RoverGridWorld problem is solved using value iteration applied to a flat MDP structure and a bi-level MDP structure. The flat MDP reasons at a granular level while the bi-level MDP consists of a high-level MDP that decides the order of visiting targets and a low-level MDP that handles path planning to the selected target. The paths chosen by the two frameworks are equivalent in terms of Manhattan distance. In both cases, the total reward accrued at the end of the traverse is the same. However, as discussed later, the bi-level MDP uses less computation time than the flat MDP, while achieving near-optimal policies.}}
    \label{fig:exp1-episodes}
\end{figure*}
For our first experiment, we construct a simple version of the RoverGridWorld problem with grid size $= 10\times 10$, a time horizon of 20 steps, and a discount factor ($\gamma$) of 0.95. The targets and obstacles are specified as shown in Figure \ref{fig:rewards-field}. The action space is simplified so that targets must only be visited (ignoring the concepts of measurement and drilling) i.e., 
\begin{equation*}
\Actions = \{\texttt{UP, DOWN, LEFT, RIGHT}\}.
\end{equation*}
The state space is similarly simplified to
\begin{equation*}
\States = \{\state\} = \{\left(x, y, t, \texttt{visited} \right)\}.
\end{equation*}
The transition $\Transitions$ and reward $\Rewards$ functions remain the same, disregarding any logic related to measurements or drilling. 

We compare the results of using four different solvers to solve this MDP. The first solver uses value iteration to solve the flat MDP. This is an exact solver that leads to the optimal solution. The second is our bi-level MDP solver which uses two value iteration solvers for the high-level and low-level MDPs respectively. The third and fourth solvers are a Q-learning solver and a SARSA solver, which are both reinforcement learning methods of developing a policy through exploration via simulation, i.e., without directly using the transition function, aka ``model-free'' methods. 

Qualitatively, the policies returned by the bi-level MDP solver and the flat MDP solver are similar, as expected. For most initial states, the bi-level MDP (solved with value iteration) and the flat MDP (also solved with value iteration) return the same policy, which, due to deterministic transitions, results in the same rover path. However, there are some initial states where the two solvers return different policies. One such example is shown in Figure \ref{fig:exp1-episodes}. In this case, the low-level value functions are optimal from the perspective of reaching the current high-level target, but not necessarily optimal from a full state perspective, i.e., the low-level solver can be myopic. Additionally, the high-level action selection is based on a conservative heuristic, leading to a different outcome than the flat MDP. This trade-off between global and recursive optimality has also been observed in other hierarchical MDP structures \cite{Dietterich2000,BakkerZivkovicEtAl2005}. Despite the two paths being different, we see that the rover picks up the same rewards. In fact, the bi-level value iteration and flat value iteration converge to a similar total reward value, when averaged across 500 simulations, as we discuss next. 

\begin{figure}[ht]
    \centering
    \includegraphics[width=0.49\textwidth]{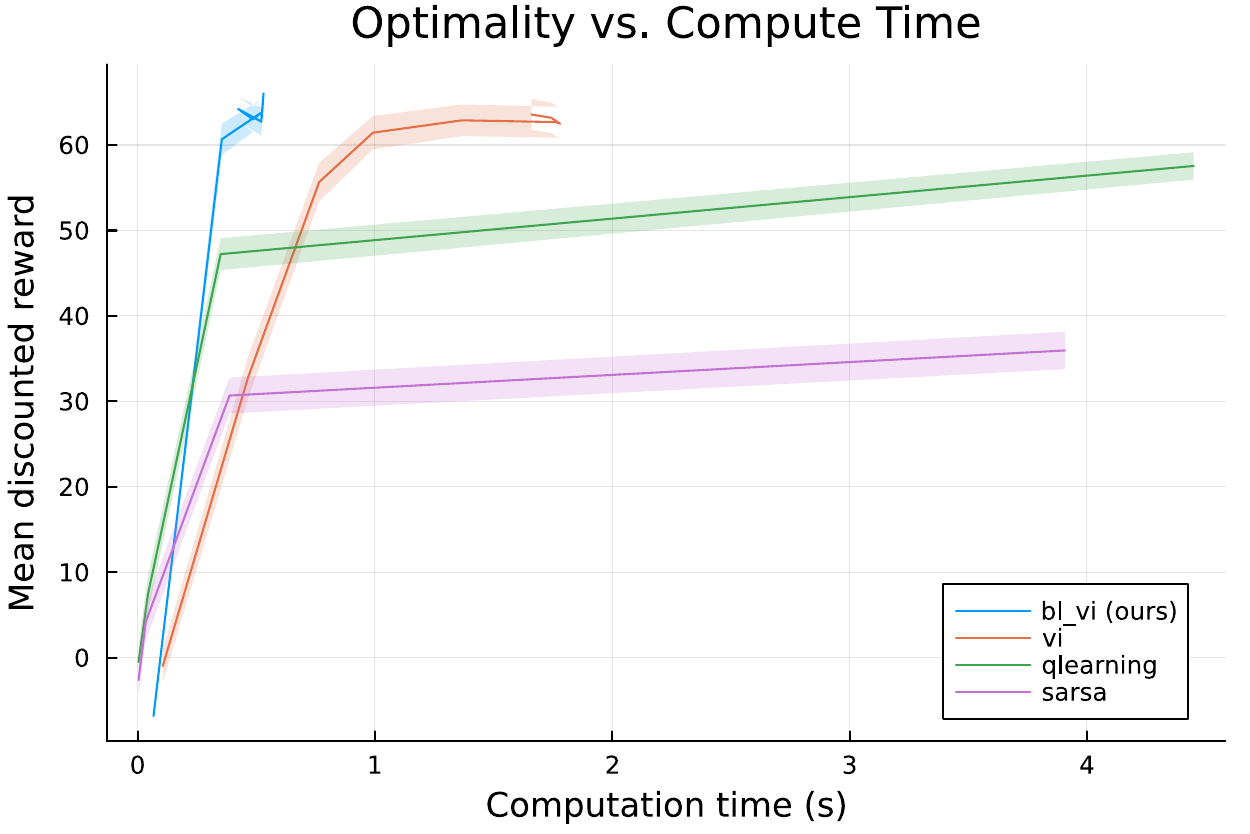}
    \caption{\bf{For experiment 1, comparing four methods of solving the RoverGridWorld problem, we see that the bi-level MDP solved with value iteration converges to the same maximum reward as the flat MDP with value iteration (which solves the problem optimally) but the compute time is about half ($\approx$0.5 vs $\approx$1 second). The reinforcement learning based solvers, Q-learning and SARSA, take longer and achieve smaller rewards, which is expected since they are model-free methods. The shading denotes the error averaged over 500 simulations of each policy. The visual oscillations at the end of the VI lines are due to running multiple runs with varying number of maximum iterations which all converge at the same iteration count, with nearly identical computation times.}}
    \label{fig:exp1-tradeoff}
\end{figure}

Quantitatively, we look at the rewards accrued by each policy (averaged over 500 simulations) in a family of policies that are generated by each solver. This family of policies is generated by varying the maximum number of iterations allowed for the solver, resulting in policies that trade-off low computation time (lower number of iterations) vs. high rewards. The results of this trade-off for all four solvers are shown in Figure \ref{fig:exp1-tradeoff}. Note that instead of plotting number of iterations, we plot the computation time because the solver iterations vary in duration and some solvers converge before hitting their maximum iteration limit. We see that the bi-level MDP formulation produces rewards that are in the same range as the flat MDP value iteration, while taking much less computation time. This shows that the bi-level MDP unlocks faster computation by leveraging the structure of the mission planning MDP to reduce the state and action space size in each level of the MDP, without sacrificing the performance of the policy. In the next experiment, we will show how this computation vs. performance trade-off changes with increasing complexity of the MDP.

\subsection{Experiment 2}
Next, we add back in the measurement and drilling actions to the action space, i.e.,
\begin{equation*}
\Actions = \{\texttt{UP, DOWN, LEFT, RIGHT, MEASURE, DRILL}\}
\end{equation*}
and the state space is also restored to the full space, i.e., 
\begin{equation*}
\States = \{\state\} = \{\left(x, y, t, \texttt{measured}, \texttt{drilled} \right)\}.
\end{equation*}
The rewards function requires that a measurement be taken near each target (in a neighboring grid cell), before the target can be drilled. The measurement yields a small reward and the drilling yields a large reward. Each target can only be measured and drilled once. The transition function is made stochastic, so that the time it takes to do a measurement or a drilling varies between 1-3 timesteps. Note that for simplicity, the time component of the state is kept discrete so that we can still compare frameworks using the discrete value iteration solver.

\begin{figure}[t]
    \centering
    \includegraphics[width=0.49\textwidth]{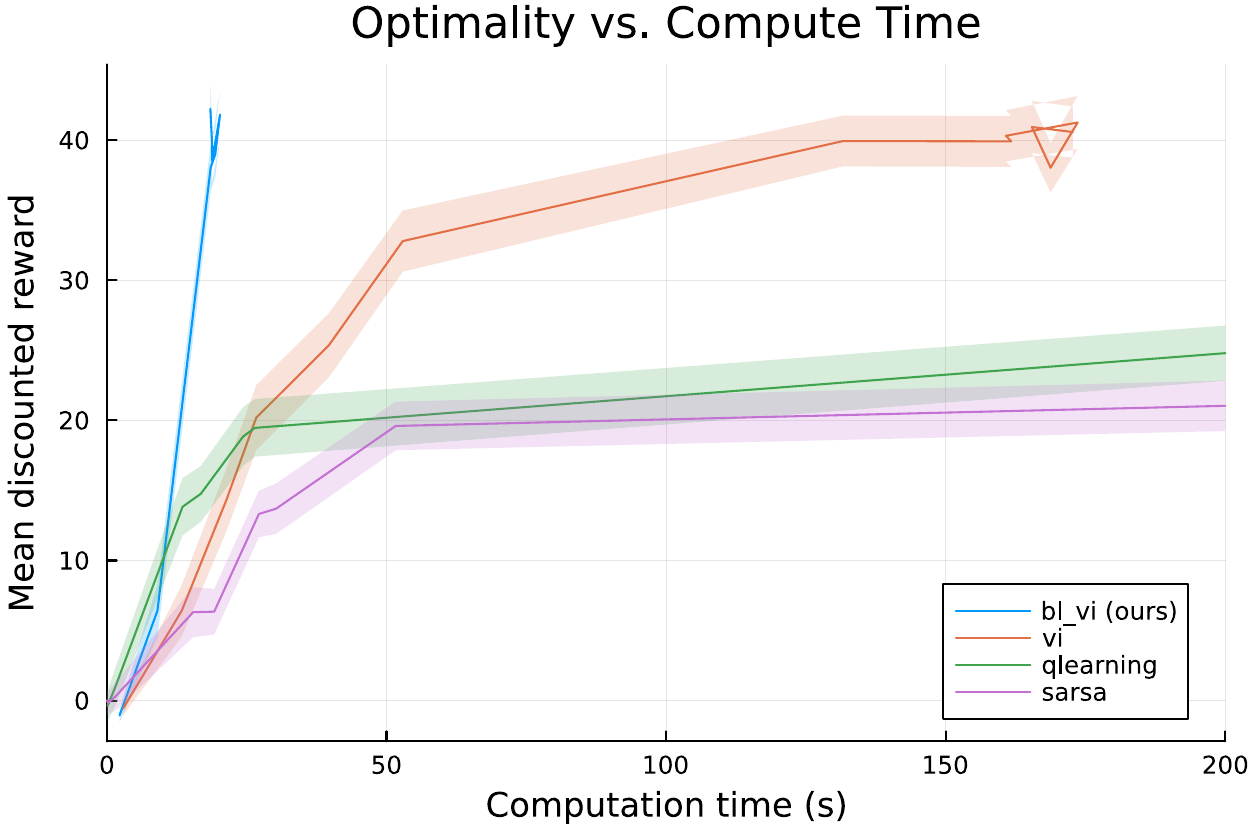}
    \caption{\bf{For experiment 2, the same four methods are compared for a larger RoverGridWorld problem. The bi-level MDP solved with value iteration converges to the optimal policy much faster ($\approx$ 20 seconds) compared to the flat MDP solved with value iteration ($>$ 100 seconds). The shading denotes the error averaged over 500 simulations of each policy. The visual oscillations at the end of the VI lines are due to running multiple runs with varying number of maximum iterations which all converge at the same iteration count, with similar computation times.}}
    \label{fig:exp2-tradeoff}
\end{figure}

Figure \ref{fig:exp2-episodes} shows the path planning solutions for a given starting point on a $10 \times 10$ grid with a time horizon of 25, and two targets to visit. We see that the final solutions for the flat MDP and the bi-level MDP are identical. A more complex problem ($50 \times 50$ grid, a time horizon of 100, 10 randomly placed targets, and gradually moving sun shadows similar to Figure \ref{fig:rewards-field}) is solved using the four solvers discussed previously and the results are shown in Figure \ref{fig:exp2-tradeoff}. Value iteration on the bi-level MDP formulation converges in $\approx 20$ seconds while the flat MDP takes $\approx 100$ seconds. This hints at the fact that that the computational advantage of the bi-level MDP formulation widens with the increased complexity (state/action space size and time horizon) of the mission planning problem. In experiment 3, we further analyze this relationship between problem complexity and computation time.

\subsection{Experiment 3}
\begin{figure}[hb!]
    \centering
    \includegraphics[width=0.49\textwidth]{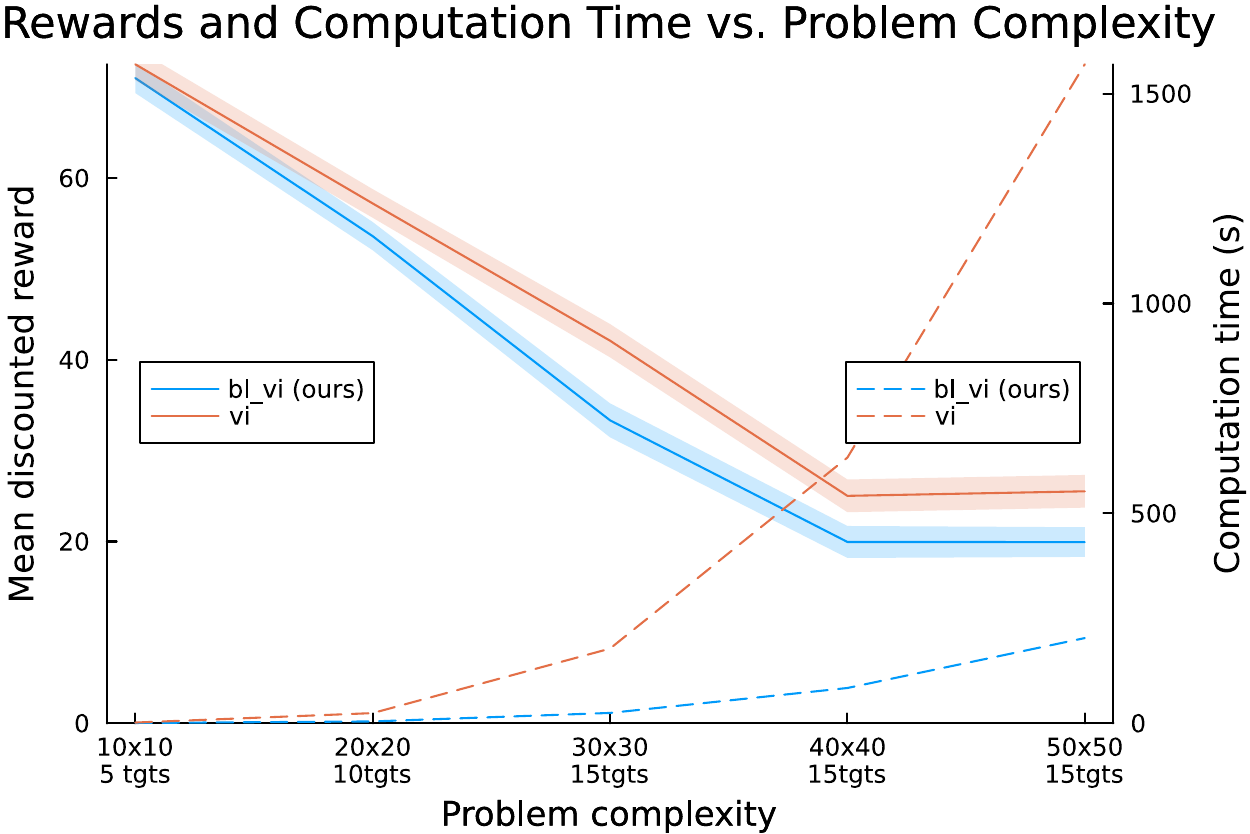}
    \caption{\bf{For experiment 3, the problem ``complexity'' is varied, by varying the grid size between $10 \times 10$ to $50 \times 50$ and placing targets at random. The solid lines show the rewards accrued and the dashed lines show the computation time. The bi-level MDP formulation {\normalfont $\texttt{bl\_vi}$} uses significantly less computation time (13\% of the flat MDP {\normalfont $\texttt{vi}$} time, for the most complex problem) while achieving near-optimal reward (lowest being 78\% of the {\normalfont $\texttt{vi}$} reward, which is optimal). The shading denotes the error averaged over 500 simulations of each policy.}}
    \label{fig:exp3}
\end{figure}

We maintain the same MDP formulation but vary the problem complexity, and solve the flat MDP and bi-level MDP to convergence using value iteration. As seen in Figure \ref{fig:exp3}, the bi-level MDP produces a computational advantage that increases with problem complexity, while achieving near-optimality of rewards accrued. In the most complex case, the bi-level MDP achieves 78\% of the optimal reward using 13\% of the computation time.

%%%%%%%%%%%%%%%%%%%%%%%%%%%%%%%%%%%%%%
\clearpage
\section{Conclusion}
\label{sec:conclusion}
In this work, we presented a framework to transform any mission planning MDP into a bi-level MDP, using a few pieces of additional information about the state and rewards structure, that likely exist in any mission planning scenario. We showed that this framework, when tested on a RoverGridWorld problem, significantly reduces computation time for solving, while retaining near-optimal policies. Lastly, we showed that the computational advantage increases as the action space and state spaces grow. This work enables fast generation of a policy that describe the sequence of actions to take from any \textit{arbitrary} state in the state space, which allows for quick traverse planning in the event of a deviation from or delay to the nominal path, i.e., the quick generation of a contingency branch.

Interesting directions of future work include further characterizing the optimality of the bi-level MDP framework, e.g., by proving convergence to recursive optimality, despite global suboptimality, in a vein similar to MAXQ \cite{Dietterich2000}. Another direction would be to analyze the computational improvement of a bi-level MDP when state and action spaces are \textit{continuous}. When the state has a discrete location component and a continuous time component, we could generalize existing piece-wise value iteration methods to a bi-level framework, and use them to solve a bi-level time-dependent MDP \cite{BoyanLittman2000}. Another interesting direction would be to add a third intermediate level to the MDP that monitors for ``safe exploration'' in MDPs \cite{MoldovanAbbeel2012}, i.e. removes unsafe actions from the low-level planner, to preserve ergodicity, and prevent catastrophic failure. Conversely, the intermediate layer could also be used to quantify and reward \textit{robustness} specifically for space missions, where robustness is not only the absence of failure scenarios but also the reliable and risk-sensitive accumulation of scientific value. This would encourage the low-level planner to find not just the optimal path to the target in nominal execution, but also trade-off for robustness of the path. 
%%%%%%%%%%%%%%%%%%%%%%%%%%%%%%%%%%%%%%

% %%%%%%%%%%%%%%%%%%%%%%%%%%%%%%%%%%%%%%%%%%%%%%%%%%%%%%%%%%%%%%%%%%%%%%%%%%%%%%%%%%%%%%%%%%%%%%%%%
% \appendices{}              % note there is no {} to put a title. Each appendix has its own title
% %%%%%%%%%%%%%%%%%%%%%%%%%%%%%%%%%%%%%%%%%%%%%%%%%%%%%%%%%%%%%%%%%%%%%%%%%%%%%%%%%%%%%%%%%%%%%%%%%
% % For a single appendix, use the \appendix{} keyword and do not use the \section command.

% \section{More Information}        % first appendix
% %%%%%%%%%%%%%%%%%%%%%%%%%%
% This is the first appendix. 

% \subsection{Comments}
% If you have only one appendix, use the ``appendix'' keyword.

% \subsection{More Comments}
% Use section and subsection keywords as usual.

% \section{Yet More Information}    % second appendix
% %%%%%%%%%%%%%%%%%%%%%%%%%%%%%%
% This is the second appendix.

%%%%%%%%%%%%%%%%%%%%%%%%%%%%%%%%%%%%%%%%%%%%%%%%%%%%%%%%%%%%%%%%%%%%%%%%%%%%%%%%%%%%%%%%%%%%%%%%%%%%%%
\acknowledgments
The authors gratefully acknowledge support from the following funding sources: Stanford Graduate Fellowship (Somrita Banerjee), The Aerospace Corporation's University Partnership Program, NASA University Leadership initiative (grant \#80NSSC20M0163), and NASA VIPER mission (Artemis Program).

%%%%%%%%%%%%%%%%%%%%%%%%%%%%%%%%%%%%%%%%%%%%%%%%%%%%%%%%%%%%%%%%%%%%%%%%%%%%%%%%%%%%%%%%%%%%%%%%%%%%%%
\bibliographystyle{IEEEtran}
\bibliography{ASL_papers,main}
% \begin{thebibliography}{1}

% \bibitem{ITAR}
% U.S. Munitions List, Sections 38 and 47(7) of the Arms Export Control Act (22 U.S.C 2778 and 2794(7).

% \bibitem{AeroConf}
% Aerospace Conference Web site: \underline{www.aeroconf.org}.

% \end{thebibliography}

%%%%%%%%%%%%%%%%%%%%%%%%%%%%%%%%%%%%%%%%%%%%%%%%%%%%%%%%%%%%%%%%%%%%%%%%%%%%%%%%%%%%%%%%%%%%%%%%%%%%%%
\thebiography
%% This biostyle allows you to insert your photo size 1in X 1.25in

\begin{biographywithpic}
{Somrita Banerjee}{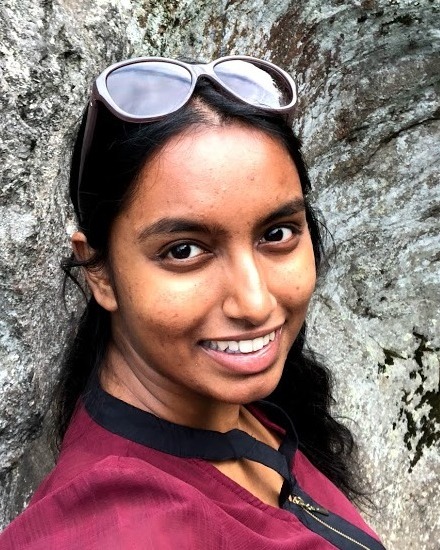}
is a Ph.D. candidate in the Autonomous Systems Lab at Stanford University. She received her B.S. in Mechanical and Aerospace Engineering from Cornell University in 2017. Her research interests lie at the intersection of trajectory optimization, machine learning, and control algorithms, specifically towards development of the next generation of space robotics.
\end{biographywithpic}

\begin{biographywithpic}
{Edward Balaban}{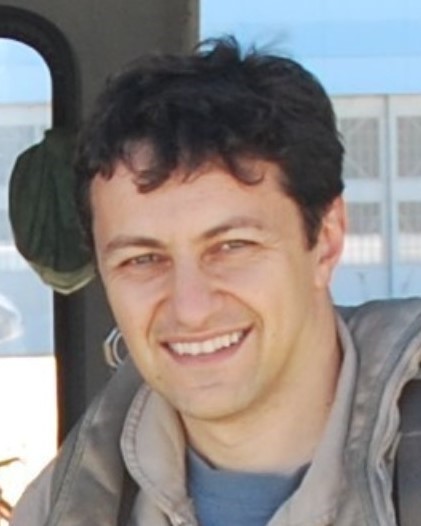}
is a scientist at NASA Ames Research Center whose professional interests include robotics, autonomy, artificial intelligence, and development of innovative space missions. He is the lead for strategic mission planning on NASA's Volatiles Investigating Polar Exploration Rover (VIPER) mission and is also a member of VIPER’s Science and Mission Operations teams. Edward holds a bachelor’s degree in Computer Science from The George Washington University, a master's degree in Electrical Engineering from Cornell University, and a Ph.D. in Aeronautics and Astronautics from Stanford University.
\end{biographywithpic}

\begin{biographywithpic}
{Mark Shirley}{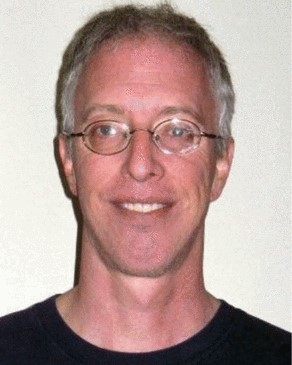}
received B.S. and M.S. degrees in Electrical Engineering and Computer Science in 1983 and a Ph.D. in Computer Science with a concentration in Artificial Intelligence from the Massachusetts Institute of Technology in 1989. He is currently a member of the research staff at NASA's Ames Research Center. His interests include diagnostic algorithms. satellite ground systems and distance learning. He previously worked at Xerox PARC and GenRad Inc.
\end{biographywithpic}

\begin{biographywithpic}
{Kevin Bradner}{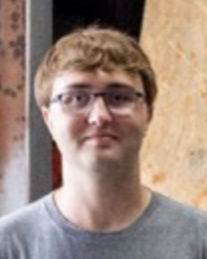}
is a computer scientist in the Data Sciences Group at NASA's Ames Research Center. He completed his M.S. in Computer Science from Case Western Reserve University in 2020. His research interests are in techniques for data-efficient robotic skill acquisition using a combination of trial-and-error learning and transfer techniques.
\end{biographywithpic}

\begin{biographywithpic}
{Marco Pavone}{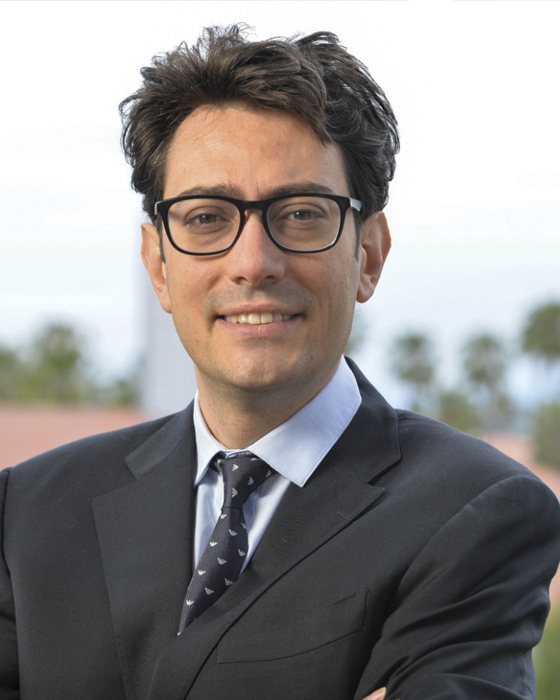}
is   an   Associate Professor of Aeronautics and Astronautics at Stanford University,  where he is the Director of the Autonomous Systems Laboratory. Before  joining  Stanford, he  was  a  Research  Technologist  within the  Robotics  Section  at  the  NASA  Jet Propulsion  Laboratory.   He  received  a Ph.D. degree in Aeronautics and Astronautics from the Massachusetts Institute of  Technology  in  2010.   His  main  research  interests  are  in the  development  of  methodologies  for  the  analysis,  design, and  control  of  autonomous  systems,  with  an  emphasis  on self-driving cars, autonomous aerospace vehicles, and future mobility systems.  He is a recipient of a number of awards, including  a  Presidential  Early  Career  Award  for  Scientists and Engineers, an ONR YIP Award, an NSF CAREER Award, and a NASA Early Career Faculty Award.  He was identified by the American Society for Engineering Education (ASEE) as one of America’s 20 most highly promising investigators under the age of 40.
\end{biographywithpic}

\end{document}